\documentclass[journal]{IEEEtran}
\usepackage[table]{xcolor}
\usepackage{graphicx,amsmath,amssymb,epstopdf,mathtools,multirow,booktabs}
\usepackage{subcaption}	
\usepackage{amsthm,float}
\usepackage{threeparttable, tablefootnote}
\usepackage{algpseudocode}
\usepackage{arydshln}
\usepackage{pdfrender}

\usepackage[hidelinks]{hyperref}
\usepackage{paralist}

\floatstyle{ruled}
\newfloat{algorithm}{tbp}{loa}
\providecommand{\algorithmname}{Algorithm}
\floatname{algorithm}{\protect\algorithmname}

\let\vec\boldvec

\newcommand{\trsp}{{\!\scriptscriptstyle\top}}

\usepackage{tikz}
\newcommand*\circled[1]{\tikz[baseline=(char.base)]{
            \node[shape=circle,draw,inner sep=0.4pt] (char) {#1};}}

\begin{document}
\title{
Auto-LfD: Towards Closing the Loop for Learning from Demonstrations}

\author{Shaokang Wu, Yijin Wang, and Yanlong Huang

\thanks{All authors are with the School of Computing, University of Leeds, Leeds LS29JT, UK. (\tt\small scswu@leeds.ac.uk; ml20y3w@leeds.ac.uk; y.l.huang@leeds.ac.uk).}
}

\maketitle

\begin{abstract}
Over the past few years, there have been numerous works towards advancing the generalization capability of robots, among which learning from demonstrations (LfD) has drawn much attention by virtue of its user-friendly and data-efficient nature. 
While many LfD solutions have been reported, a key question has not been properly addressed: how can we evaluate the generalization performance of LfD? For instance, when a robot draws a letter that needs to pass through new desired points, how does it ensure the new trajectory maintains a similar shape to the demonstration? 
This question becomes more relevant when a new task is significantly far from the demonstrated region. To tackle this issue, a user often resorts to manual tuning of the hyperparameters of an LfD approach until a satisfactory trajectory is attained. In this paper, we aim to provide closed-loop evaluative feedback for LfD and optimize LfD in an automatic fashion.
Specifically, we consider dynamical movement primitives (DMP) and kernelized movement primitives (KMP) as examples and develop a generic optimization framework capable of measuring the generalization performance of DMP and KMP and auto-optimizing their hyperparameters without any human 
inputs. Evaluations including a peg-in-hole task and a pushing task on a real robot evidence the applicability of our framework.

\end{abstract}

\section{Introduction}

Learning from demonstrations (LfD), as a long-standing research topic in the scope of robot learning, aims to endow robots with the capability of mimicking human behaviours from a single or a few pre-provided demonstrations \cite{atkeson1997robot}. Notably, additional policy updates via interaction with the environment \cite{stulp2013robot} or guidance from human users \cite{celemin2022interactive} can be incorporated into a LfD paradigm. 

There is a large body of literature in LfD that has been dedicated to addressing two key problems: (\emph{i}) what to learn and (\emph{ii}) how to learn. For the first problem, various works (e.g., \cite{zeestraten2017approach,abu2021probabilistic,saveriano2023learning}) studied the learning of different forms of demonstrations depending on task requirements, e.g., Cartesian position and velocity, orientation and angular velocity, joint position and velocity, stiffness and damping matrices, manipulability, interaction forces, as well as various combinations of these profiles. 
For the second problem, many LfD solutions have been proposed under different assumptions about the learning model. For instance, by encoding a demonstration as a spring-damper model with an additional forcing term modulating the acceleration profile, dynamical movement primitives (DMP) were developed \cite{ijspeert2013dynamical}. Task-parameterized Gaussian mixture model (GMM) was proposed in \cite{calinon2016tutorial}, where consistent features underlying demonstrations within each task frame were modelled by GMM and later these features, together with new task frames, were utilized to deal with task generalization. In \cite{paraschos2013probabilistic}, a demonstration was assumed to be the weighted sum of a series of basis functions and multivariate Gaussian distribution was used to model the distribution of multiple demonstrations, leading to a probabilistic LfD framework, i.e., probabilistic movement primitives (ProMP). In contrast to ProMP which maximizes the likelihood of demonstrations, we considered the posterior instead and introduced a non-parametric LfD solution \cite{KMP}, i.e., kernelized movement primitives (KMP), where the explicit definition of basis functions was mitigated. 

While the community witnesses remarkable progress in LfD in terms of algorithms and applications, one crucial question seems to be overlooked: how to evaluate the generalization performance of LfD? Indeed, in supervised learning and reinforcement learning settings, the evaluation of generalization is straightforward, e.g., using mean squared error (MSE) for regression, cross-entropy loss (CEL) for classification, and reward for reinforcement learning. However, the design of a sensible metric for LfD is nontrivial.  

Considering the task of writing a 2-D letter and assuming that only a single demonstration of this letter is available, if we use DMP to write it from a new start-point towards a new end-point, how can we ensure the new trajectory is smooth enough and meanwhile maintains the shape of the demonstration? Note that the adapted trajectory must be different from the demonstration as a consequence of the new start-point and end-point demands. In this case, the traditional treatment \cite{fanger2016gaussian} of using MSE or maximum likelihood estimation (MLE) as a generalization metric could become problematic, especially when the adapted trajectory is far from the demonstration. Often, 
we rely on manual tuning of the hyperparameters of DMP until a proper trajectory meeting our requirements is obtained. The tuning process heavily relies on our experience and usually involves many trials, which restrains the deployment of DMP in dynamic environments since such tuning is required whenever a new task requirement arises. A similar issue of tuning hyperparameters is also encountered in other LfD solutions, including ProMP and KMP.

In this paper, we propose an automatic optimization framework for LfD (i.e., \emph{auto-LfD}) so as to free users from the laborious tuning of hyperparameters, where
a novel metric capable of measuring the generalization performance of any LfD approach is designed. Specifically,
we take the parametric method DMP and the nonparametric method KMP as examples and use this metric to guide the optimization of their hyperparameters. We begin with preliminaries on DMP and KMP in Section~\ref{sec:pre}. After that, we discuss the limitations of MSE and MLE acting as generalization metrics for LfD (Section~\ref{sec:moti}). In Section~\ref{sec:autolfd}, we present a novel metric for evaluating the generalization performance of LfD and explain hyperparameters optimization for  DMP and KMP using the metric, where both gradient descent (GD) and Bayesian optimization (BO) are exploited. We test the auto-LfD framework (Section~\ref{sec:exp}) in several scenarios including simulated writing tasks, as well as peg-in-hole and pushing tasks implemented in a real robot. We conclude this work in Section~\ref{sec:con}.

\section{Preliminaries} \label{sec:pre}
In this section, we briefly review the basic rationale of DMP (Section~\ref{subsec:dmp}) and KMP (Section~\ref{subsec:kmp}), which will be later optimized using the proposed auto-LfD framework.

\subsection{DMP} \label{subsec:dmp}
DMP, consisting of a first-order canonical system and a second-order transformation system, can learn and generalize the motion pattern underlying a single demonstration. Formally, DMP encodes trajectories as \cite{ijspeert2013dynamical}
\begin{equation}
\begin{aligned}
&\tau \dot{s} = - \alpha s, \\
&\tau^2 \Ddot{\vec{\xi}} = \vec{K}_p (\vec{g} - \vec{\xi}) - \tau \vec{K}_v \dot{\vec{\xi}}_n + s (\vec{g} - \vec{\xi}_0) \odot \vec{f}_w(s) ,
\end{aligned}
\label{equ:dmp}
\end{equation}
where $\vec{K}_p$ and $\vec{K}_v$ are stiffness and damping matrices. $\tau$, $\alpha$, and $s$ denote motion duration, decay factor, and phase variable, respectively. $\vec{\xi} \in \mathbb{R}^\mathcal{O}$ represents $\mathcal{O}$-dimensional position while $\dot{\vec{\xi}}$ and $\ddot{\vec{\xi}}$ respectively denote the corresponding velocity and acceleration. $\vec{\xi}_0$ is the starting point and $\vec{g}$ is the target. $\odot$ stands for the element-wise product. $\vec{f}_w(s) \in \mathbb{R}^\mathcal{O}$ represents the forcing term driven by $s$, where the parameter vector $\vec{w}$ is learned from the demonstration. 

In contrast to the classical representation of approximating the forcing term $\vec{f}(s)$ as a linear combination of a set of predefined basis functions, Gaussian process (GP) 
was suggested in \cite{fanger2016gaussian} to model $\vec{f}(s)$, where the explicit definition of basis functions is mitigated and fewer open parameters are demanded. 

Suppose we have access to a demonstration of time-length $N$, i.e., $\{t_n, \vec{\xi}_n,\dot{\vec{\xi}}_n,\ddot{\vec{\xi}}_n\}_{n=1}^N$, we can substitute the demonstration into (\ref{equ:dmp}) and extract a new training dataset $\{s_n, \vec{f}(s_n)\}_{n=1}^N$. The new dataset can be learned by GP to predict  
the corresponding forcing term $\vec{f}(s)$ for an arbitrary $s\in(0,1]$. 
Given an inquiry $s^*$, we have \cite{rasmussen2006gaussian} 
\begin{equation}
{f_i(s^*)}=\vec{k}^{*}( \vec{K} + \lambda \vec{I} )^{-1} \vec{U}_i,
\label{equ:gp}
\end{equation}
with 
\begin{equation*}
\begin{aligned}
&\vec{k}^{*} = [k(s^*, s_1) \ k(s^*, s_2) \ \cdots \ k(s^*, s_N)], \\
&\vec{K}_{i,j} = k(s_i, s_j),\\
&\vec{U}_i = [{f}_i(s_1) \ {f}_i(s_2) \ \cdots \  {f}_i(s_N)]^\trsp,
\end{aligned}
\end{equation*}
where $k(\cdot,\cdot)$ denotes a kernel function, e.g., the definition of a commonly used squared exponential (SE) kernel is $k(s_i,s_j)=\exp({-k_h(s_i-s_j)^2})$ with a hyperparameter $k_h>0$. $\vec{k}^* \in \mathbb{R}^{1\times N}$, $\vec{U}_i \in \mathbb{R}^{N}$, $\vec{K}_{i,j}$ denotes the element at the $i$-th row and the $j$-th column of the matrix $\vec{K} \in \mathbb{R}^{N\times N}$.
${f}_i(\cdot)$ corresponds to the $i$-th element of $\vec{f}(\cdot)$. $\lambda>0$ is a scalar, and $\vec{I}$ is an identity matrix.

\subsection{KMP} \label{subsec:kmp}

Unlike DMP which learns a single demonstration, KMP learns the probabilistic distribution of multiple demonstrations. Given $H$ demonstrations $\{\{t_{n, h}, \vec{\xi}_{n, h}, \dot{\vec{\xi}}_{n, h} \}_{n=1}^N\}_{h=1}^H$ with $\vec{\xi}_{n,h}$ being the trajectory point at the $n$-th time step from the $h$-th demonstration, their distribution can be modelled by GMM and Gaussian mixture regression (GMR) \cite{calinon2016tutorial,GMR}, leading to a probabilistic reference trajectory that encapsulates the distribution of demonstrations, i.e., 
$\{t_n, \hat{\vec{\mu}}_n, \hat{\vec{\Sigma}}_n \}_{n=1}^N$,
where $\mathcal{P}\bigl( \left[\begin{matrix}
    \vec{\xi}_n \\ \dot{\vec{\xi}}_n 
\end{matrix}\right]|t_n \bigr)=\mathcal{N}(\hat{\vec{\mu}}_n,\hat{\vec{\Sigma}}_n)$.

Let us denote $\hat{\vec{\mu}} = [{\hat{\vec{\mu}}_1}^\trsp \ {\hat{\vec{\mu}}_2}^\trsp \ \cdots \ {\hat{\vec{\mu}}_N}^\trsp]^\trsp$ and $\hat{\vec{\Sigma}}= \text{blockdiag}(\hat{\vec{\Sigma}}_1, \hat{\vec{\Sigma}}_2,\ldots,\hat{\vec{\Sigma}}_N)$. For a query input $t^{*}$, KMP predicts its corresponding output as \cite{KMP}
\begin{equation}
    \vec{\mu}(t^{*}) = \left[\begin{matrix}
    \vec{\xi}(t^{*}) \\ \dot{\vec{\xi}}(t^{*}) 
\end{matrix}\right] = \vec{k}^{*}( \vec{K} +\lambda \hat{\vec{\Sigma}} )^{-1} \hat{\vec{\mu}},
    \label{equ:kmp}
\end{equation}
where
\begin{equation*}
\begin{aligned}
&\vec{k}^{*}= \left[\vec{k}(t^{*}, t_1) \ \vec{k}(t^{*}, t_2) \ \cdots \ \vec{k}(t^{*}, t_N)\right], \\
&\vec{K}_{i,j} = \vec{k}(t_i, t_j).
\end{aligned}
\end{equation*}
Both $\vec{k}^{*}$ and $\vec{K}$ depend on the extended kernel matrix $\vec{k} (\cdot,\cdot)$ whose elements are ${k}(\cdot,\cdot)$ and the associated first-order and second-order derivatives, see \cite{KMP} for more details. 
$\lambda>0$ is used to mitigate the overfitting issue. 

Note that the predictions in (\ref{equ:gp}) and (\ref{equ:kmp}) have a nonparametric form, and proper hyperparameter tuning is needed, 
including the kernel parameter $k_h$ involved in the kernel function $k(\cdot,\cdot)$ and the regularization factor $\lambda$. Throughout the paper, we will use the notation $\vec{\Theta} = [k_h \ \lambda]^\trsp$ to represent the collection of hyperparameters to be optimized.

\section{Motivation} \label{sec:moti}
Why do we need a novel metric, rather than MSE and MLE, to measure the generalization performance of LfD? In order to answer this question,  
we first formulate the generalization (i.e., adaptation) problem in LfD (Section~\ref{subsec:problem}), and subsequently present some examples to evidence the issues arising from the (weighted) distance-aware metrics MSE and MLE (Section~\ref{subsec:lim}).

\subsection{Problem formulation \label{subsec:problem}}

For the sake of brevity, we rewrite the demonstration of DMP as $\hat{\vec{\mu}} = [{\hat{\vec{\mu}}_1}^\trsp \ {\hat{\vec{\mu}}_2}^\trsp \ \cdots \ {\hat{\vec{\mu}}_N}^\trsp]^\trsp$, where $\hat{\vec{\mu}}_n = [\vec{\xi}_n^\trsp \ \dot{\vec{\xi}}_n^\trsp]^\trsp$. 
Note that the same notation $\hat{\vec{\mu}}_n$ is also used to denote the mean of the reference trajectory in KMP, which should be straightforwardly distinguished from the context. Given a demonstration $\{t_n, \hat{\vec{\mu}}_n\}_{n=1}^N$ for DMP or a probabilistic reference trajectory $\{t_n, \hat{\vec{\mu}}_n, \hat{\vec{\Sigma}}_n \}_{n=1}^N$ for KMP, as well as adaption constraints $\vec{c}$ (e.g., desired positions of new start-point, via-point, and end-point), we can generate an adapted trajectory as $\vec{\mu}^* = {[{\vec{\mu}_1^{*\trsp}} \  \vec{\mu}_2^{*\trsp} \ \cdots \ \vec{\mu}_N^{*\trsp}]}^\trsp$ via DMP or KMP, where each predicted datapoint $\vec{\mu}_n^*=[\vec{\xi}_n^{*\trsp} \, \dot{\vec{\xi}}_n^{*\trsp} ]^\trsp \in \mathbb{R}^{2\mathcal{O}}$ comprises both position and velocity.

\begin{figure*}[bt]
    \centering
        \includegraphics[width=1\linewidth]{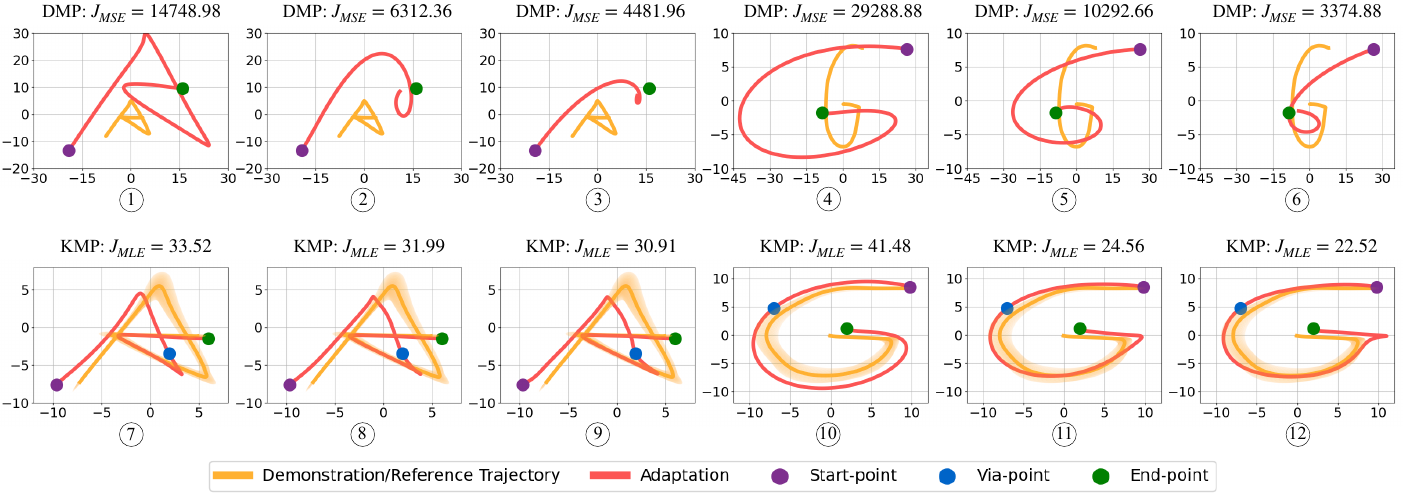}
    \caption{Evaluations of the MSE metric on DMP (\emph{top row}) and the MLE metric on KMP (\emph{bottom row}). \emph{Top row}: in \textcircled{1}--\textcircled{3}, DMP is used to adapt the 2-D demonstration `A' towards a new start point and a new target point, where each adaptation corresponds to a different set of hyperparameters of DMP; similar applications of DMP in writing the letter `G' are showcased in \textcircled{4}--\textcircled{6}.
    \emph{Bottom row}: KMP learns the probabilistic reference trajectory of the 2-D letter `A' or `G', and generates adapted trajectories towards a new start-point, via-point, and end-point with different hyperparameters.}
    \label{fig:mse&mle}
\end{figure*}

\begin{figure}[bt]
    \centering
        \includegraphics[width=0.99\linewidth]{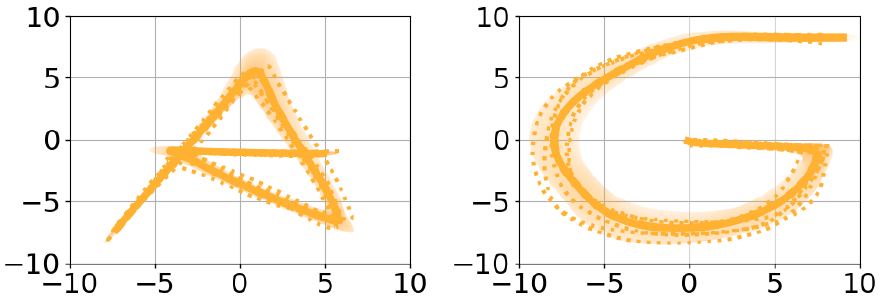}
    \caption{Extracting the probabilistic reference trajectory from multiple demonstrations using GMR, where the yellow dotted curves denote demonstrations. The yellow solid curves and the shaded areas depict the means and covariances of the probabilistic reference trajectories, respectively.}
    \label{fig:gmr}
\end{figure}

Our goal is to build an automatic optimization framework (which is referred to as auto-LfD) for DMP and KMP to guide the optimization of their hyperparameters $\vec{\Theta}$. To do so, a metric  $J(\hat{\vec{\mu}}, \vec{\mu}^*)$ that measures the discrepancy between the demonstration $\hat{\vec{\mu}}$ and the adapted trajectory $\vec{\mu}^*$ will be required. Moreover, such a metric should take trajectory smoothness into account. Suppose we have such a metric $J(\cdot,\cdot)$ at hand, we can view trajectory adaptation as a function $\vec{\mu}^* =\mathcal{G}_{\vec{\Theta}}(\hat{\vec{\mu}},\vec{c})$ and formulate the hyperparameter optimization for LfD as minimizing  $J(\hat{\vec{\mu}}, \vec{\mu}^*) =  J (\hat{\vec{\mu}}, \mathcal{G}_{\vec{\Theta}} ( \hat{\vec{\mu}}, \vec{c}))$.

\subsection{Limitations of MSE and MLE \label{subsec:lim}}
We use MSE as the generalization metric for DMP and use MLE for KMP since DMP learns a single demonstration and KMP essentially maximizes the observation probability of multiple demonstrations. The MSE and MLE metrics are
\begin{equation}
\begin{aligned}
J_{\textrm{MSE}}( \hat{\vec{\mu}},\vec{\mu}^*) &= \frac{1}{N} \sum_{n=1}^N (\vec{\mu}_{n}^* - \hat{\vec{\mu}}_n)^{\trsp} (\vec{\mu}_{n}^* - \hat{\vec{\mu}}_n),  \\
J_{\textrm{MLE}}(\hat{\vec{\mu}},\vec{\mu}^*) &= \frac{1}{N} \sum_{n=1}^N (\vec{\mu}_{n}^* - \hat{\vec{\mu}}_n)^{\trsp} \hat{\vec{\Sigma}}_n^{-1} (\vec{\mu}_{n}^* - \hat{\vec{\mu}}_n).
\end{aligned}
\label{equ:loss}
\end{equation}

To illustrate the limitations of MSE and MLE, we report their applications in writing 2-D letters `A' and `G'. The first and second rows of Fig.~\ref{fig:mse&mle} respectively show the adapted trajectories using DMP and KMP, where for either method different hyperparameters are tested. In the second row of Fig.~\ref{fig:mse&mle}, the probabilistic reference trajectories for `A' and `G' are extracted from their corresponding five demonstrations via GMR, as shown in Fig.~\ref{fig:gmr}.
 
In the plots \circled{1}--\circled{3} of Fig.~\ref{fig:mse&mle}, the adapted trajectories (shown as red curves) for the letter `A' become distorted when the MSE cost decreases, showing that the MSE cost fails to indicate the shape maintenance. Similarly, MSE is unable to measure the adaptation performance of DMP in writing the letter `G' (see plots \circled{4}--\circled{6}), where the adapted trajectory that resembles the shape of the demonstration and reaches the new target precisely, however, has the largest MSE cost (see \circled{4}).
 
In the second row of Fig.~\ref{fig:mse&mle}, we evaluate the MLE costs on the trajectories generated by KMP, where a desired via-point is imposed for either letter in addition to the desired new start-point and end-point. As with the observations in the first row, the best adaptations (plotted by red curves) have the largest MLE costs (see \circled{7} for `A' and \circled{\footnotesize10} for `G'), while the adapted trajectories with significant distortions have smaller MLE costs, see \circled{9} and \circled{\footnotesize12}. 
 
From the above examples, we can conclude that optimizing the hyperparameters of DMP with the MSE metric and KMP with the MLE metric could be problematic, since both metrics focus on the `reproduction' of the demonstration or the reference trajectory (i.e., staying close to the demonstration or the reference trajectory in terms of (weighted) Euclidean distance) without considering the motion shape and smoothness requirements, thus failing to provide a reliable indicator of the generalization performance of LfD methods. 

It is worth emphasizing that similar limitations of MSE and MLE apply to other distance-aware metrics as well. For example, the Fr\'echet distances \cite{hertel2021similarity} corresponding to the plots \circled{1} and \circled{3} in Fig.~\ref{fig:mse&mle}
are 224.74 and 165.18, respectively, while the Fr\'echet distances of the plots \circled{4} and \circled{6} are 401.38 and 101.82, respectively, showing that the Fr\'echet distance is also not a proper indicator of adaptation performance.

\begin{figure*}[bt]
    \centering
\includegraphics[width=0.68\textwidth]{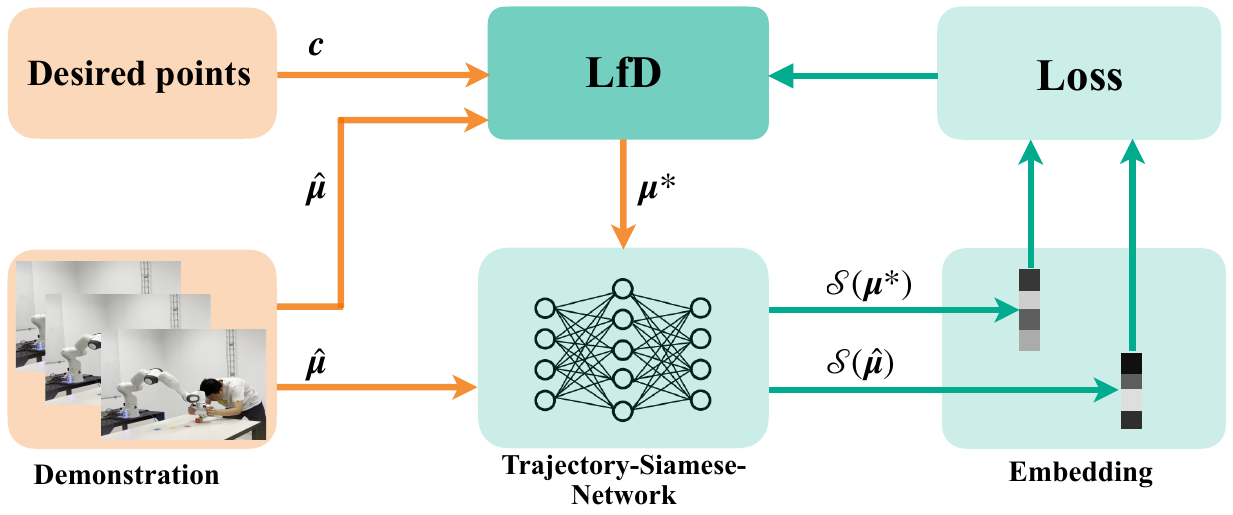}
    \caption{An overview of the auto-LfD framework, where the hyperparameters of an LfD approach are constantly optimized towards reducing the distance between the feature vectors $\mathcal{S}(\hat{\vec{\mu}})$ and $\mathcal{S}({\vec{\mu}}^*)$. Here, $\hat{\vec{\mu}}$ and $\vec{\mu}^*$ represent the demonstration and the adapted trajectory generated by LfD, respectively.}
    \label{fig:framework}
\end{figure*}

\section{Auto-LfD} \label{sec:autolfd}
We have discussed the issues of using MSE and MLE as adaptation metrics, now we propose a novel framework auto-LfD that allows for optimizing the hyperparameters of DMP and KMP automatically, where an encoder network that transforms trajectories into a latent feature space is designed (Section~\ref{subsec:enc}). With the encoder network, the generalization metric can be designed naturally (Section~\ref{subsec:metric}). After that, we discuss two routes to hyperparameter optimization, including GD (Sections~\ref{subsec:gd}) and BO (Section~\ref{subsec:bo}).

\subsection{Trajectory Siamese network \label{subsec:enc}}

Siamese networks \cite{bromley1993signature,chopra2005learning} have been proven effective in many applications, e.g., face recognition and signature verification. 
The Siamese network amounts to an encoder mapping inputs to feature vectors that are subsequently used to calculate the similarity between inputs. In order to train a Siamese network suitable for measuring the adaptation performance of LfD, we collect a training dataset consisting of $M$ triplets $<\hat{\vec{\mu}}_m,\vec{p}_m,\vec{n}_m>$, where $m\in\{1, 2, \ldots, M\}$, $\vec{p}_m$ and $\vec{n}_m$ respectively represent satisfactory (i.e., `positive' samples) and unsatisfactory (i.e., `negative' samples) adapted trajectories in comparison with the demonstration $\hat{\vec{\mu}}_m$. Note that the input of the Siamese network is a trajectory comprising $N$ trajectory points (each point is a concatenation of position and velocity) rather than a single trajectory point. 

Formally, we train the Siamese network with the triplet loss \cite{tripletloss} by minimizing
\begin{equation}
\sum_{m=1}^M \max \bigl(0, \gamma + || \mathcal{S}(\hat{\vec{\mu}}_m) - \mathcal{S}(\vec{p}_{m}) ||_2 -
  ||\mathcal{S}(\hat{\vec{\mu}}_m) - \mathcal{S}(\vec{n}_{m}) ||_2  \bigr) ,
    \label{equ:triplet_loss}
\end{equation}
where $\mathcal{S}(\cdot):\mathbb{R}^{2\mathcal{O}N} \rightarrow \mathbb{R}^{h}$ corresponds to the encoding of a trajectory into an $h$-dimensional feature vector, $\lVert \cdot \rVert_2$ represents $\ell_2$ norm.
The constant margin $\gamma>0$ ensures that the positive and negative samples can be more easily distinguished.

\subsection{An automatic optimization framework for LfD \label{subsec:opti}}

\subsubsection{Generalization metric \label{subsec:metric}}
Following the spirit of the loss function in (\ref{equ:triplet_loss}) -- `positive' trajectories should stay close to the demonstration while `negative' trajectories should stay away from the demonstration, we can define a metric to measure the distance (i.e., dissimilarity) between a demonstration $\hat{\vec{\mu}}$ and an adapted trajectory $\vec{\mu}^*$ under condition $\vec{c}$, i.e.,
\begin{equation}
\mathcal{L}_{\vec{\Theta}}
=  ||\mathcal{S}(\hat{\vec{\mu}}) - \mathcal{S}(\vec{\mu}^*) ||_2
=||\mathcal{S}(\hat{\vec{\mu}}) - \mathcal{S} 
(\mathcal{G}_{\vec{\Theta}}(\hat{\vec{\mu}}, \vec{c}) )||_2 .
\label{equ:cost}
\end{equation}
With the optimal hyperparameters $\vec{\Theta}$ that minimizes $\mathcal{L}_{\vec{\Theta}}$, both DMP and KMP are able to generate adapted trajectories that mostly resemble the demonstration. Note that the metric in (\ref{equ:cost}) operates the level of trajectories, which can be combined with many other LfD methods beyond DMP and KMP.

\subsubsection{Gradient descent \label{subsec:gd}}
In order to search for the optimal $\Vec{\Theta}$ minimizing $\mathcal{L}_{\vec{\Theta}}$ in (\ref{equ:cost}), a common optimization technique is a gradient-based method, i.e., GD. Specifically, GD iteratively updates $\vec{\Theta}$ via $\vec{\Theta} := \vec{\Theta} - \eta \nabla_{\vec{\Theta}} \mathcal{L}_{\vec{\Theta}}$, where $\eta>0$ is the learning rate. $\nabla_{\vec{\Theta}} \mathcal{L}_{\vec{\Theta}}$ stands for the gradient of (\ref{equ:cost}) with respect to $\vec{\Theta}$ and is computed by chain rule, i.e.,
\begin{equation}
    \nabla_{\vec{\Theta}} \mathcal{L}_{\vec{\Theta}} = \nabla_{\mathcal{S}(\vec{\mu}^*)} \mathcal{L} \bigl(\mathcal{S}(\hat{\vec{\mu}}), \mathcal{S}(\vec{\mu}^*)\bigr) \nabla_{\vec{\mu}^*} \mathcal{S}(\vec{\mu}^*)  \nabla_{\vec{\Theta}} \mathcal{G}_{\vec{\Theta}}(\hat{\vec{\mu}},\vec{c}),
\label{equ:backpropagation_vector}
\end{equation}
where 
\begin{equation*}
\begin{aligned}
    &\nabla_{\mathcal{S}(\vec{\mu}^*)} \mathcal{L} (\mathcal{S}(\hat{\vec{\mu}}), \mathcal{S}(\vec{\mu}^*)) = \frac{ \mathcal{S}(\vec{\mu}^*) - \mathcal{S}(\hat{\vec{\mu}}) }{\lVert \mathcal{S}(\hat{\vec{\mu}}) - \mathcal{S} ( \vec{\mu}^*) \rVert_2}, \\
&\vec{\mu}^* = \mathcal{G}_{\vec{\Theta}}(\hat{\vec{\mu}}, \vec{c}).
\end{aligned}
\end{equation*}
The computation of $\nabla_{\vec{\Theta}} \mathcal{G}_{\vec{\Theta}}(\hat{\vec{\mu}}, \vec{c})$ in (\ref{equ:backpropagation_vector}) is straightforward in KMP as each predicted trajectory point only depends on $\vec{\Theta}$. 
In contrast, an analytical expression of this gradient is complicated in DMP since DMP uses an `accumulated' way to generate trajectories, i.e., a current trajectory point depends on all previous points. However, such a derivative can be automatically determined in a modern deep learning framework, e.g., Pytorch.

\subsubsection{Bayesian optimization \label{subsec:bo}}
In contrast to GD which relies on explicit calculation of gradient and could converge to a local minimum, BO is a gradient-free, global optimization method. BO resorts to an acquisition function to decide the query points for evaluations towards finding the optimal input point with a minimal number of searching steps, where the core idea is to balance the exploitation and exploration when sampling queries. 

We take one of the most popular acquisition functions expected improvement (EI) \cite{frazier2018tutorial} as an example. Suppose we have evaluated a set of hyperparameters 
$\{\vec{\Theta}_1,\vec{\Theta}_2, \ldots, \vec{\Theta}_n\}$ and obtained their corresponding metric costs $\{\mathcal{L}(\vec{\Theta}_1),\mathcal{L}(\vec{\Theta}_2), \ldots, \mathcal{L}(\vec{\Theta}_n)\}$ via (\ref{equ:cost}), the next query point $\vec{\Theta}_{n+1}$ is determined by
\begin{equation}
\begin{aligned}
\vec{\Theta}_{n+1} &= \mathop{\mathrm{argmax}}_{\vec{\Theta}}
\mathbb{E}_{ \mathcal{L} (\vec{\Theta})}\max\bigl(\mathcal{L}^* - \mathcal{L} (\vec{\Theta}), 0\bigr).
\end{aligned}
\label{equ:EI}
\end{equation}
Here, the expectation is estimated over the distribution $\mathcal{L} (\vec{\Theta}) \sim \mathcal{GP}(\mu_{\mathcal{L}|\mathcal{D}}, K_{\mathcal{L}|\mathcal{D}})$, which is predicted by GP at the input $\vec{\Theta}$ given observations $\mathcal{D}=\{\vec{\Theta}_i,\mathcal{L}(\vec{\Theta}_i)\}_{i=1}^n$. $\mathcal{L}^*$ is the smallest one among $\{\mathcal{L}(\vec{\Theta}_i)\}_{i=1}^n$. Once the new query point $\vec{\Theta}_{n+1}$ is known and evaluated using (\ref{equ:cost}), a new pair $\{\vec{\Theta}_{n+1},\mathcal{L}(\vec{\Theta}_{n+1})\}$ will be added to the existing observations $\mathcal{D}$, and subsequently the search strategy in  (\ref{equ:EI}) will be employed again to decide the following query point $\vec{\Theta}_{n+2}$. By repeating the above procedure, the optimal $\vec{\Theta}^*$ associated with the minimal metric loss $\mathcal{L}(\vec{\Theta}^*)$ will be found.

An overview that summarizes the proposed auto-LfD framework is provided in Fig.~\ref{fig:framework}. Given a demonstration (or a reference trajectory) $\hat{\vec{\mu}}$, adaptation constraints $\vec{c}$, and hyperparameters $\vec{\Theta}$, we can use DMP or KMP to generate an adapted trajectory $\vec{\mu}^*$ that addresses the imitation of the demonstration and the adaptation constraints (i.e., desired points). Subsequently, we can measure the dissimilarity between the demonstration and the adapted trajectory by comparing their feature vectors $\mathcal{S}(\hat{\vec{\mu}})$ and $\mathcal{S}({\vec{\mu}}^*)$ (encoded by the Siamese network) using the metric in (\ref{equ:cost}). The hyperparameters of DMP or KMP can be optimized via GD or BO towards reducing the metric loss (\ref{equ:cost}). Once the optimization converges, the adapted trajectory from DMP or KMP using the optimal hyperparameters $\vec{\Theta}^*$ will be a proper generalization from the demonstration $\hat{\vec{\mu}}$ under the constraints $\vec{c}$.

\begin{figure*}[bt]
    \centering
    \centering
\includegraphics[width=0.85\linewidth]{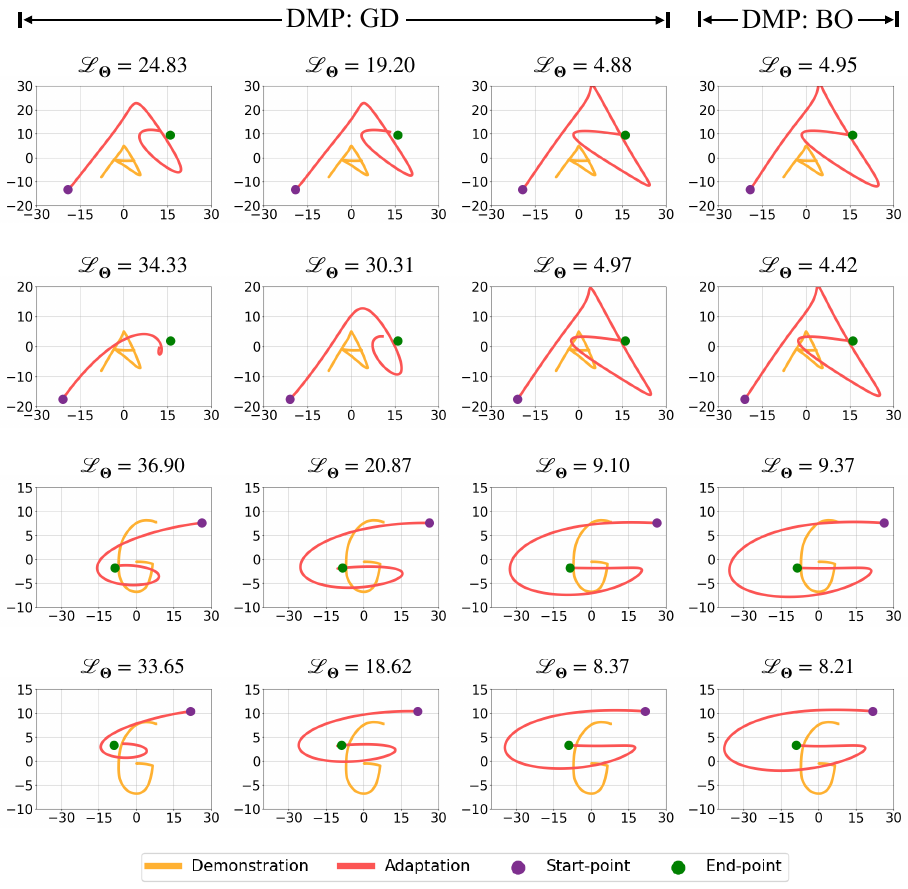}
    \caption{Evaluations of auto-DMP in letter-writing tasks. The first, second and third columns correspond to the updated hyperparameters of DMP via GD, where the adapted trajectories are gradually improved in terms of reaching the desired end-point and resembling the shape of the demonstration. The fourth column shows the adapted trajectories with the optimal hyperparameters found by BO.} 
    \label{fig:auto-dmp}
\end{figure*}

\section{Evaluations} \label{sec:exp}

In this section, we report the evaluations of auto-LfD in simulated letter-writing tasks (Section~\ref{subsec:eva:letter}), as well as peg-in-hole and pushing tasks in real robotic settings (Section~\ref{subsec:eva:real}). Specifically, 
we answer the following questions in the evaluations:
\begin{itemize}
    \item[(i)] Is the metric in (\ref{equ:cost}) a proper indicator of the generalization performance? 
    \item[(ii)] Which optimization algorithm between GD and BO can achieve superior performance? 
    \item[(iii)] Can auto-LfD generate satisfactory adapted trajectories? 
\end{itemize}

\subsection{Letter-writing tasks \label{subsec:eva:letter}}

Table~\ref{table:sia} lists the relevant parameters involved in training the Siamese encoder network. 
We employ GD and BO to optimize the hyperparameters of DMP and KMP towards reducing the metric cost in (\ref{equ:cost}), respectively. The Mat\'ern kernel is used in BO.

\subsubsection{Optimizing the hyperparameters of DMP \label{subsec:dmp:simu}} We test the writing of letters `A' and `G' using our framework and DMP (i.e., \emph{auto-DMP}), where the letters are obtained from \cite{calinon2017learning}.
The adapted trajectories for both letters are plotted in Fig.~\ref{fig:auto-dmp}. Take the first row as an example, the adapted trajectory (depicted by the red curve in the first plot) generated by DMP with an initial setting of hyperparameters fails to reach the desired end-point and maintain the shape of the demonstration (plotted by the yellow curve). After updating the hyperparameters $\vec{\Theta}$ of DMP via GD a few times, the adapted trajectory is improved in terms of the trajectory shape (see the second plot), but it is still unable to reach the desired target precisely. After 30 updates, the final trajectory (see the third figure) reaches the desired target while resembling the shape of the demonstration. Note that careful initialization of hyperparameters for GD is demanded in all evaluations in Fig.~\ref{fig:auto-dmp} since GD could be trapped into an inappropriate local minimum.

In Fig.~\ref{fig:auto-dmp}, the desired points (i.e., start-point and end-point) in the first row are the same as the ones in \circled{1}--\circled{3} in Fig.~\ref{fig:mse&mle}, and the desired points in the third row are the same as the ones in \circled{4}--\circled{6} in Fig.~\ref{fig:mse&mle}. However, differing from the MSE metric, the proposed metric loss $\mathcal{L}_{\vec{\Theta}}$ in (\ref{equ:cost}) indeed decreases when the adapted trajectory becomes better. Similarly, in the second and fourth rows, the smaller the metric loss is, the better the adaptation is.

The adapted trajectories under the optimized hyperparameters via BO are plotted in Fig.~\ref{fig:auto-dmp}, showing that BO can achieve satisfactory trajectories for different letters and adaptation conditions. Here, proper initialization is not required for BO since BO is essentially a sampling-based method that uses the acquisition function to direct the sampling process.

\begin{figure*}[bt]
    \centering
    \centering
    \includegraphics[width=0.85\linewidth]{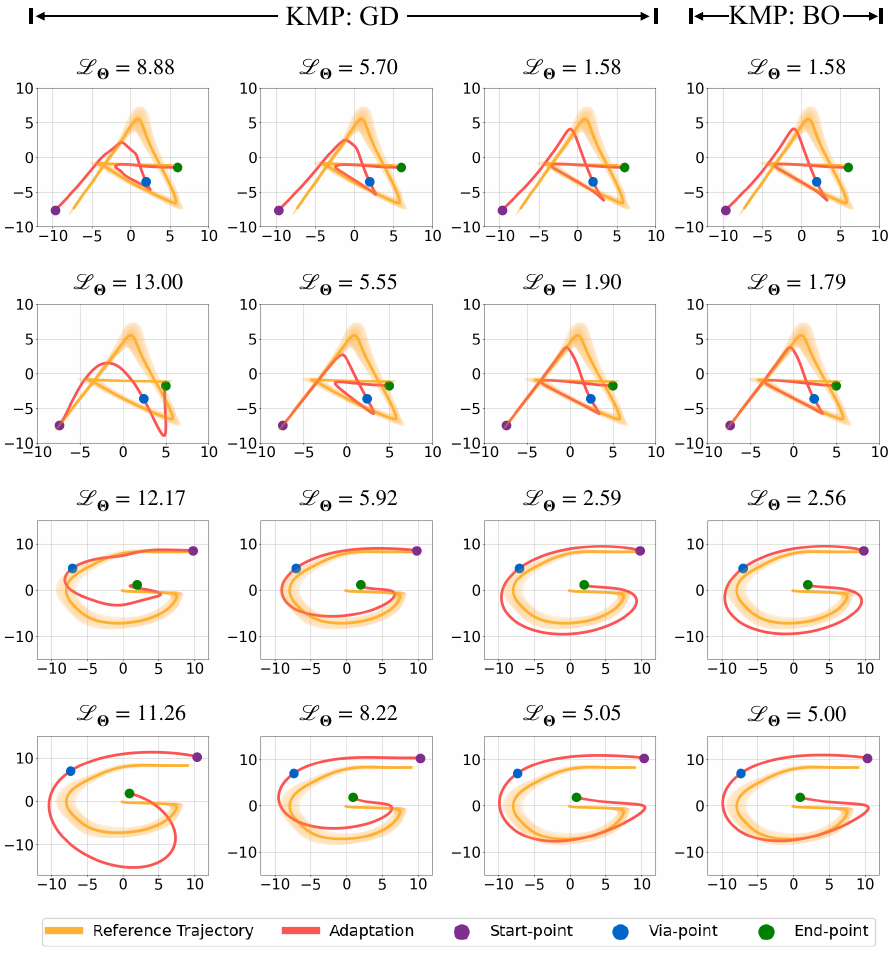}
    \caption{Evaluations of auto-KMP on letter-writing tasks, where both GD and BO are implemented. The probabilistic reference trajectory for either letter is extracted from five demonstrations using GMM and GMR.}
    \label{fig:auto-kmp}
\end{figure*}

\begin{figure*}[bt]
    \centering    
    \includegraphics[width=1\textwidth]{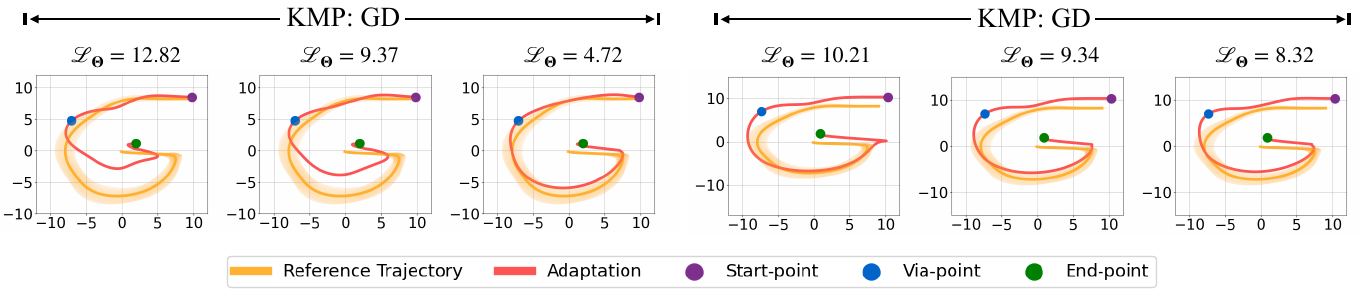}
     \caption{Evaluations of auto-KMP with GD optimization, where the hyperparameters of KMP are initialized with inappropriate values.}
    \label{fig:gd-init}
\end{figure*}

\subsubsection{Optimizing the hyperparameters of KMP \label{subsec:kmp:simu}}
In addition to the evaluations on DMP, we assess the performance of auto-LfD on KMP (i.e., \emph{auto-KMP}) as well, see Fig.~\ref{fig:auto-kmp}. Note that the first and third rows of Fig.~\ref{fig:auto-kmp} have the same desired points as \circled{7}--\circled{9} and \circled{10}--\circled{12} in Fig.~\ref{fig:mse&mle}, respectively. Unlike MLE, our metric can indicate the adaptation performance properly, i.e., the generalization using KMP improves as the metric loss decreases. Moreover, by using either GD or BO, our framework can optimize the hyperparameters of KMP towards better generalization under different adaptation conditions.

\begin{table}[bt]
	\vspace{0.55cm}
      \centering
      \caption {Training Parameters of the Siamese Network in Writing Tasks}
		\begin{tabular}{p{3cm} c p{3cm}}
			\toprule
			\textbf{Hyperparameters} & \textbf{Values} \\   
                \midrule
                Learning rate & $10^{-7}$ \\
                Batch Size & 200 \\
                Epoch & 30000 \\
                Margin & 0.5 \\
                Data Size & 3026 \\
			\bottomrule
	   \end{tabular}
	\label{table:sia}
\end{table}

\subsubsection{Comparison between GD and BO \label{subse:gd:bo}}
While GD and BO yield similar performances in Fig.~\ref{fig:auto-dmp} and Fig.~\ref{fig:auto-kmp}, there is an essential difference between them. In fact, GD is sensitive to the initial values of the hyperparameters. Given inappropriate initialization, GD could lead to undesired convergence, see examples of utilizing auto-KMP in Fig.~\ref{fig:gd-init}. Although the first three columns and the last three columns of Fig.~\ref{fig:gd-init} have the same desired points as the third and fourth rows of Fig.~\ref{fig:auto-kmp}, the adapted trajectories in Fig.~\ref{fig:gd-init} either reach the desired end-point with an abrupt change and (or) exhibit certain distortion. Compared with GD which may lead to undesired local minimum, BO is a global optimization technique that provides a more stable solution for the auto-LfD framework. Therefore, we only adopt BO for the following evaluations in real robotic tasks. 

Note that in the first and third rows of Fig.~\ref{fig:auto-dmp}, the metric costs $\mathcal{L}_{\vec{\Theta}}$ from BO (i.e., 4.95 for `A' and 9.37 for `G') are larger than the optimal costs from GD (i.e., 4.88 for `A' and 9.10 for `G') by neglectable margins, given that the margins only take a tiny proportion of the initial costs (i.e., 24.83 and 36.90). In fact, if we sample more samples in BO (100 iterations are used in Fig.~\ref{fig:auto-dmp}), BO can lead to smaller costs, but the improvement of trajectories will be hardly observed for both letters.

\subsection{Real robot experiments \label{subsec:eva:real}}

So far, we have reported the performance of auto-LfD in writing 2-D letters under various constraints of desired points, we now carry out real-world experiments using a seven-degree-of-freedom robot. Specifically, we assess the efficacy of the auto-LfD framework by performing a peg-in-hole task using DMP and a pushing task using KMP.
Table \ref{table:real:siamese} summarizes the settings of training Siamese networks in both tasks.

\begin{table}[bt]
	\vspace{0.6cm}
	\caption {Training Parameters of the Siamese Network in Real Robotic Tasks}
	\centering
		\begin{tabular}{lccc}
			\toprule
			\textbf{Hyperparameters} & \textbf{Peg-in-Hole Task} & \textbf{Pushing Task} \\   
                \midrule
                Learning rate & $10^{-5}$ & $10^{-5}$ \\
                Batch Size & 100 & 50 \\
                Epoch & 20000 & 20000 \\
                Margin & 0.5 &0.5 \\
                Data Size & 431 & 301 \\
			\bottomrule
	   \end{tabular}
	\label{table:real:siamese}
\end{table}

\subsubsection{Peg-in-hole task}

\begin{figure*}[bt]
    \centering
        \centering
    \includegraphics[width=0.99\linewidth]{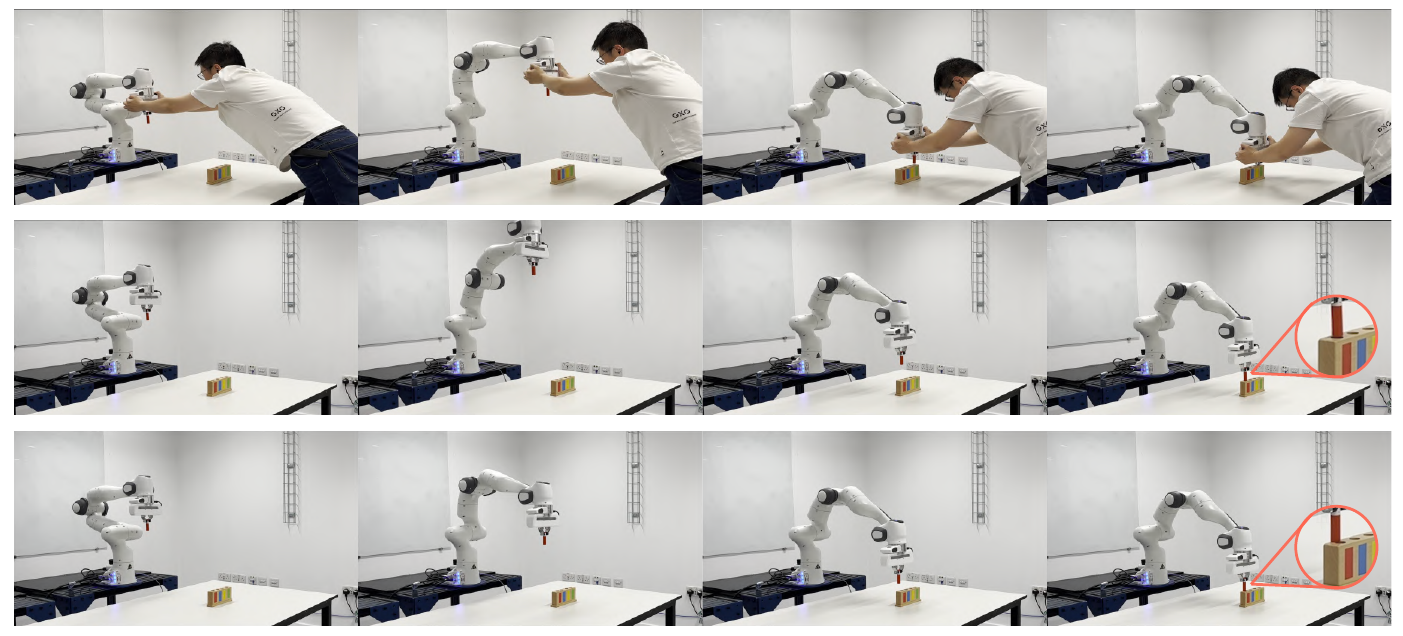}
    \caption{Snapshots of the peg-in-hole task. \emph{First row} shows the kinesthetic teaching of the peg-in-hole task. \emph{Second and third rows} correspond to the adapted robot trajectories that are optimized using our metric in (\ref{equ:cost}) and the MSE metric.}
    \label{fig:snapshot:peg}
\end{figure*}

The goal of the peg-in-hole task is to insert a peg into a desired hole. An illustration of collecting a demonstration in such a task is shown in the first row of Fig.~\ref{fig:snapshot:peg}. Given that the diameter of the hole is slightly larger than that of the peg, it is crucial that the robot inserts the peg into the hole from a vertically downward direction when the peg is approaching the hole. Any deviation from a vertical insertion motion could result in misalignment between the peg and the hole, ultimately leading to unsuccessful tasks. Since this task demands a start-point (i.e., the initial position of the peg) and an end-point (i.e., the location of the hole), we employ auto-DMP to generalize the demonstration to unseen new tasks.

\begin{figure}[bt]
    \centering
        \centering    \includegraphics[width=0.99\linewidth]{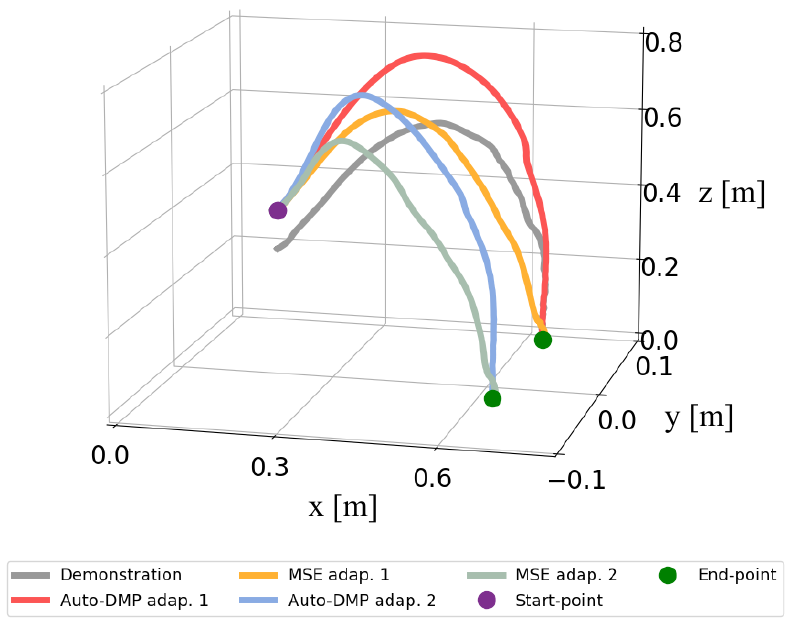}
    \caption{The demonstration and adapted real robotic trajectories in the peg-in-hole task, where two adaptation settings are considered and in either setting our metric (i.e., auto-DMP) and the MSE metric are implemented, respectively.}
    \label{fig:peg-in-hole}
\end{figure}

We collect a single demonstration for the peg-in-hole task, as depicted by the grey curve in Fig.~\ref{fig:peg-in-hole}. To verify the effectiveness of auto-DMP, 
we consider two settings: (\emph{i}) a new start-point (see the purple dot in Fig.~\ref{fig:peg-in-hole}) that is away from the initial point of the demonstration and an end-point that is the same as that of the demonstration; (\emph{ii}) a new start-point that is the same as the one used in (\emph{i}) and a new end-point that is far from the target of the demonstration. As a comparison, we also implement hyperparameter updates for DMP using the MSE metric and BO in both settings.

The real robotic trajectories in both evaluation settings are plotted in Fig.~\ref{fig:peg-in-hole}, where we can see that the adapted trajectories with our metric exhibit vertical insertion motion near the desired end-points while the trajectories with the MSE metric approach the end-points from oblique directions. Snapshots of the experiments in the second evaluation setting are provided in Fig.~\ref{fig:snapshot:peg}, where the robot using the MSE metric indeed fails to insert the peg into the desired hole (see the third row of Fig.~\ref{fig:snapshot:peg}) as a consequence of the collision between the peg and the rim of the hole. In contrast, the robot can accomplish the peg-in-hole task successfully using auto-DMP (see the second row of Fig.~\ref{fig:snapshot:peg}).

\begin{figure*}[bt]
    \centering
    \begin{minipage}{\linewidth}
        \centering
    \includegraphics[width=0.99\linewidth]{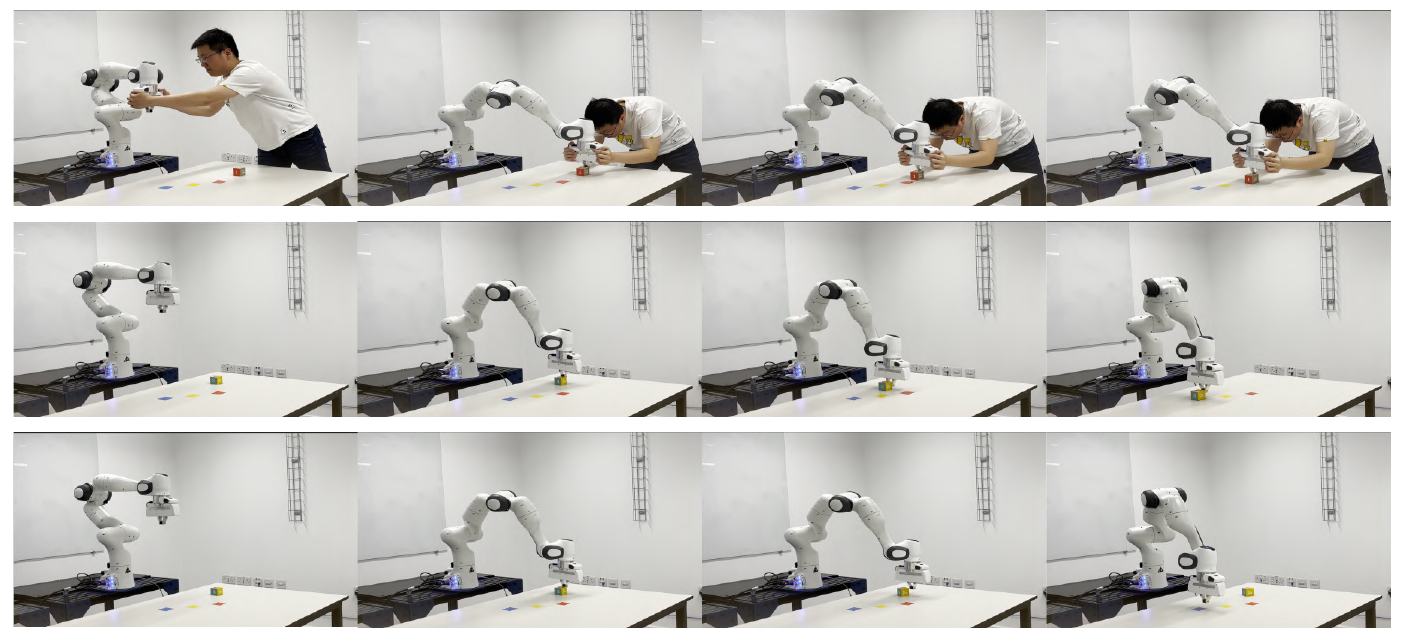}
    \end{minipage}
    \caption{Snapshots of the pushing task. \emph{First row} illustrates the process of collecting a demonstration. \emph{Second and third rows} show the robot executing the adapted trajectories from optimizing our metric and the MLE metric, respectively.}
    \label{fig:snapshot:push}
\end{figure*}

\begin{figure*}[bt]
    \centering
        \centering    \includegraphics[width=0.921\linewidth]{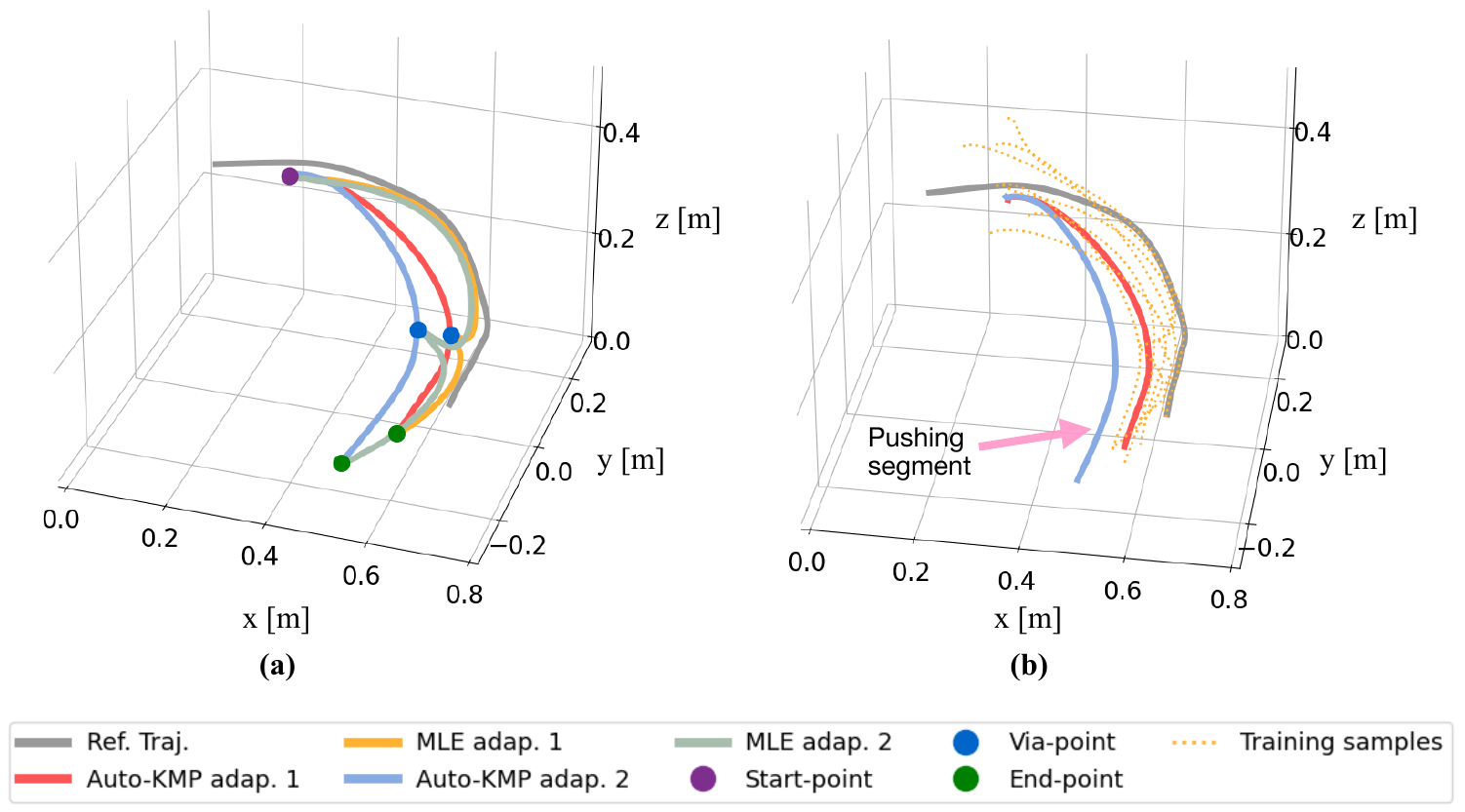}
    \caption{The evaluations of auto-KMP and the MLE metric in the pushing task. 
    (\emph{a}) shows two groups of evaluations and in either group both our metric (i.e., auto-KMP) and the MLE metric are employed in hyperparameter optimization, respectively; (\emph{b}) plots some representative samples for training the Siamese network, while the remaining samples stay within the space formed by these samples.}
    \label{fig:pushing}
\end{figure*}

\subsubsection{Pushing task} 
The pushing task involves two subtasks: reaching the small block at a desired location and pushing it towards a desired target. We can solve such a task by setting three desired points: a start-point describing the initial state of the robot's gripper, a via-point specifying the location of the block, and an end-point defining the target. In contrast to DMP, KMP provides a straightforward way to incorporate a desired via-point, so we implement KMP within our framework (i.e., auto-KMP) to accomplish the pushing task.

The procedure of collecting a demonstration is illustrated in the first row of Fig.~\ref{fig:snapshot:push}. We collect five demonstrations for the pushing task and subsequently use GMM and GMR to extract a probabilistic reference trajectory, depicted by the grey curve in Fig.~\ref{fig:pushing}(\emph{a}). We consider two settings for adaptation evaluations and both require new start-, via-, and end-points that are away from the reference trajectory. In addition to auto-KMP, we study the performance of the MLE metric as a baseline.
The adapted robotic trajectories are plotted in Fig.~\ref{fig:pushing}(\emph{a}), where the trajectories (plotted by the yellow and green curves) optimized with the MLE metric pass through different desired points precisely, whereas the trajectory shapes have significant distortions around the desired via-points. In contrast, the trajectories (plotted by the red and blue curves) generated by auto-KMP go through the desired points while keeping the shape of the reference trajectory. 

The experimental snapshots, corresponding to the second evaluation scenario in Fig.~\ref{fig:pushing}(\emph{a}), are given in Fig.~\ref{fig:snapshot:push}. In the second row of Fig.~\ref{fig:snapshot:push}, the robot equipped with auto-KMP can push the block from a new desired via-point to a new desired end-point successfully. In the third row of Fig.~\ref{fig:snapshot:push}, using the MLE metric the robot can first reach the block but soon lose physical contact when the robot bypasses the block (see the distortions in Fig.~\ref{fig:pushing}(\emph{a}) as well), thus failing to push the block towards the target. For more experimental details of the peg-in-hole and pushing tasks, please refer to the video in the supplementary material.

We emphasize that the pushing segment of the second adaptation (i.e., the blue curve in Fig.~\ref{fig:pushing}(\emph{b})) lies beyond the region covered by the dataset used for training the Siamese encoder network. For the sake of clear observation, we plot some representative training samples in Fig.~\ref{fig:pushing}(\emph{b}), while the remaining samples are confined within the space spanned by these samples. Thus, auto-KMP shows an extrapolation capability, allowing for reliable generalization outside the region of the training dataset for the Siamese network.

\section{Conclusions \label{sec:con}} 
In this paper, we have introduced a closed-loop framework auto-LfD allowing for optimizing the hyperparameters of LfD in an automatic manner, where a novel metric that measures the generalization performance of LfD is developed. Unlike the traditional MSE and MLE metrics, our metric acts as a reliable indicator when evaluating task adaptations.  The performance of auto-LfD has been verified on DMP and KMP through various tasks, including the writing, peg-in-hole and pushing tasks. 

While we focus on trajectory adaptation in Euclidean space, further extensions of this work could address more complex scenarios, including geometry-aware skills (such as orientation \cite{zeestraten2017approach} and stiffness matrix \cite{abu2021probabilistic}) and constrained skill learning \cite{huang2020linearly}.

\bibliographystyle{IEEEtran}
\bibliography{bibiography}

\end{document}